%% file: main.tex
\documentclass[conference]{IEEEtran}
\IEEEoverridecommandlockouts
\usepackage{cite}
\usepackage{amsmath,amssymb,amsfonts}
\usepackage{algorithmic}
\usepackage{graphicx}
\usepackage{textcomp}
\usepackage{csquotes}

\usepackage[table,xcdraw]{xcolor}

\usepackage{subcaption}
\usepackage{algorithmic}
\usepackage{algorithm}%

\usepackage{color}
\usepackage{comment}

\usepackage{color, colortbl}

\usepackage[hyphens]{url}
\usepackage{multirow}

\definecolor{Gray}{gray}{0.9}
\definecolor{LightCyan}{rgb}{0.88,1,1}

\begin{document}
\IEEEoverridecommandlockouts

\title{Crowdsensing-based Road Damage Detection Challenge (CRDDC’2022)}

\author{
\IEEEauthorblockN{
	Deeksha Arya${}^\star$\thanks{${}^\star$deeksha@ct.iitr.ac.in}\IEEEauthorrefmark{5},
	Hiroya Maeda\IEEEauthorrefmark{3},
	Sanjay Kumar Ghosh\IEEEauthorrefmark{4},
	Durga Toshniwal\IEEEauthorrefmark{2},
    Hiroshi Omata\IEEEauthorrefmark{5},\\
    Takehiro Kashiyama\IEEEauthorrefmark{8},
    Yoshihide Sekimoto\IEEEauthorrefmark{5}\\
    }
\IEEEauthorblockA{
\\
    \IEEEauthorrefmark{5} \textit{Centre for Spatial Information Science, The University of Tokyo, 4-6-1 Komaba, Tokyo, Japan} \\
    \IEEEauthorrefmark{3} \textit{UrbanX Technologies, Inc., Tokyo, Japan} \\
	\IEEEauthorrefmark{4} \textit{Department of Civil Engineering, Indian Institute of Technology Roorkee, 247667, India} \\
	\IEEEauthorrefmark{2} \textit{Department of Computer Science and Engineering, Indian Institute of Technology Roorkee, 247667, India} \\
	\IEEEauthorrefmark{8} \textit{Faculty of Economics, Osaka University of Economics, Japan} \\
	}
}

\maketitle

\input{abstract}
\input{introduction}

\input{ChallengeOverview}
\input{Submission_and_Evaluation}
\input{Results_and_Ranking}
\input{Discussion_and_FutureScope}
\input{conclusion}
\input{acknowledgment}

\bibliographystyle{IEEEtran}
\bibliography{references}

\end{document}

%% file: abstract.tex
\begin{abstract}
This paper summarizes the Crowdsensing-based Road Damage Detection Challenge (CRDDC), a Big Data Cup organized as a part of the IEEE International Conference on Big Data'2022. The Big Data Cup challenges involve a released dataset and a well-defined problem with clear evaluation metrics. The challenges run on a data competition platform that maintains a real-time online evaluation system for the participants. In the presented case, the data constitute 47,420 road images collected from India, Japan, the Czech Republic, Norway, the United States, and China to propose methods for automatically detecting road damages in these countries. More than 60 teams from 19 countries registered for this competition. The submitted solutions were evaluated using five leaderboards based on performance for unseen test images from the aforementioned six countries. This paper encapsulates the top 11 solutions proposed by these teams. The best-performing model utilizes ensemble learning based on YOLO and Faster-RCNN series models to yield an F1 score of 76\% for test data combined from all 6 countries. The paper concludes with a comparison of current and past challenges and provides direction for the future.

\end{abstract}

%% file: introduction.tex
\section{Introduction}

Road safety continues to be a pressing concern, with road traffic accidents being a global disaster hampering the life of millions \cite{haghani2022road, tavakkoli2022evidence, shah2022road}. Poor road conditions are one of the major factors leading to road accidents \cite{bhattacharya2022application, jakobsen2022influence}. Apropos, there is an immediate need for fresh inventions and an upward technological trajectory to reform the road and transportation arena. The paper addresses the domain of AI (Artificial Intelligence)-driven road condition inspection \cite{bhattacharya2022application} through the Crowdsensing-based Road Damage Detection challenge $(CRDDC$’2022). The background information for CRDDC is provided as follows.\\
A lack of financial resources makes many local governments unable to conduct sufficient road condition inspections on time \cite{arya2021damage}. Some municipalities automate road damage detection by using high-performance sensors. Nevertheless, the high cost of these sensors makes it infeasible to use them at the country level owing to the vast area of roads to be inspected.
This leads to the need of a system that makes it easy to assess road conditions and identify the damage to the road surface at a low cost. Recognizing this need, the first Road Damage Detection Challenge, RDDC’2018, was organized as an IEEE Big Data Cup in 2018 to evaluate the contemporary methods working towards the same goal\cite{ale2018road, alfarrarjeh2018deep, kluger2018region, mandal2018automated, manikandan2018varying, pham2018cvexplorer, wang2018deep, wang2018road}. After the challenge, several municipalities in Japan started utilizing the proposed automatic road damage detection system \cite{maeda2020}. The practical use and the feedback of government agencies suggested that the algorithms need to be more robust. Further, it was observed that most of the existing models are limited to road conditions in a single country. The development of a method that applies to more than one country leads to the possibility of designing a stand-alone system for road damage detection worldwide. \\
Considering the requirement, Arya et al.\cite{arya2021damage} augmented the Japanese dataset with road damage images from India and the Czech Republic. The authors proposed the data, Road Damage Dataset-2020 (RDD2020 \cite{rdd2020, rdd2020datasets}), which comprises 26620 images, almost thrice the volume of the 2018 dataset\cite{maeda2018road} utilized for $RDDC'2018$. Further, the authors\cite{arya2021damage} conducted a comprehensive analysis of models trained using different combinations of the data.

This RDD2020 data was made publicly available and formed the basis for organizing the second challenge of the road damage detection challenge series, named the Global Road Damage Detection Challenge, GRDDC'2020 \cite{GRDDC2020}. It invited the participants to propose models capable of efficiently detecting road damage in India, Japan, and the Czech Republic. GRDDC received an overwhelming response from the research community with a participation of 121 teams(\cite{2020rank1,rdd2020rank2, 2020rank3, 2020rank4, 2020rank5, 2020rank6, 2020rank7, 2020rank8, 2020rank9_hascoet2020fasterrcnn}).

Considering the wide impact of RDDC’2018 and GRDDC'2020, the challenge CRDDC has been brought forward in 2022. For RDDC’2018 and GRDDC'2020, the dataset was provided solely by the challenge organizers,  and participants were restricted to using only the challenge dataset to train the models. However, the CRDDC’2022, allows the participants to develop/propose their datasets, eliminating the constraint of RDDC’2018 and GRDDC’2020. Thus, the central theme of the challenge is \textquote {Crowdsensing,} – which refers to the crowdsourcing of sensor data collected as per the choice of contributors. 
It may be noted that, even though CRDDC allowed participant to propose their own dataset, it is different than the data-centric challenge organized by Behzadian et al.\cite{behzadian2022_Data_Science_for_pavemnet_challenge}. Behzadian et al. asked the participants to increase the accuracy of a predefined model architecture by utilizing various data modification methods such as cleaning, labeling and augmentation. The current challenge CRDDC'2022 is different in the following ways:

\begin{enumerate}
    \item The model architecture is not predefined. Participants may propose any architecture.
    \item The data modification methods such as cleaning and augmentation are allowed to increase the accuracy of the proposed model in Phase 3 of the challenge. However, adding new data or labels is allowed only in Phase 1 of the challenge.
\end{enumerate}

Anyone having the dataset could register for CRDDC as a data contributor, data recommender, or information provider by participating in the first phase of the challenge. Phase 2 involved a suitability analysis to select the datasets for releasing through the CRDDC'2022, and utilization for Phase 3 tasks. This paper presents a summary of this challenge, along with the solutions proposed by the participants during various phases.

%% file: ChallengeOverview.tex
\section{Challenge Review}

\subsection{Overview}

The CRDD challenge is designed to push the state-of-the-art in detecting road damages forward. The challenge comprises two components: (i) a publicly available dataset of road images and annotation and (ii) an online competition and workshop. The CRDDC'2022 dataset consists of annotated road damage images collected from India, Japan, the Czech Republic, Norway, the United States, and China. 
There are two key challenges: classification – \textquote{does the image contain any instances of a particular road damage class (where the road damage classes include longitudinal cracks, transverse cracks, potholes, etc.)?}, and detection – \textquote{where are the instances of a particular road damage class in the image (if any)?}
The data challenges are issued with deadlines each year, and a workshop is held to compare and discuss that year's results and methods. The datasets and associated annotation and evaluation methods are subsequently released and made available for use at any time.

The procedure for the previous challenge, GRDDC'2020, was inspired by the PASCAL VOC challenge that was organized from 2005 to 2012 (see \cite{everingham2010pascal, everingham2015pascal}). The procedure for CRDDC’2022 is similar to GRDDC; however, it includes some extra parts like the data contributed by the participants, new leaderboards, etc. Also, like GRDDC, the objectives of the CRDDC are twofold: first to provide challenging road damage images and high-quality annotation, together with a standard evaluation methodology to compare the performance of algorithms fairly (the dataset component); and second, to measure state of the art for the current year (the competition component).
This paper aims to describe the challenge, the methods used, the evaluation carried out, and the results. The comparative analysis of three challenges conducted till 2022 is also covered.

\subsection{Dataset}
The data released through CRDDC is named RDD2022 and is publicly available \cite{RDD2022_data}. It constitutes road images collected from India, Japan, the Czech Republic, Norway, the United States, and China. Our companion paper \cite{RDD2022_paper} presents the details of the data, including the data collection methodology, the study area, the annotation procedure, etc. These details are not repeated here.
The challenge divides the data in two chunks: Train and Test. The set \textquote{Train} consists of road images with annotations in XML files in PASCAL VOC format. The \textquote{Test} set is released without the annotations for evaluating the solutions proposed by the participants. The participants need to predict the annotation for the test images using the proposed model. The predicted annotations are evaluated by an online server accessible on the challenge website. The evaluation procedure is described in the next section.
The statistics for the distribution of images in the six countries are provided in figure 1. The training data contains annotation with marked labels for four types of road damage. Figure 2 shows the number of instances for each damage type in training data. The reason for considering these four damage types is the same as provided in \cite{arya2021damage}.

\begin{figure}[!ht]
\centering
\includegraphics[width=1.0\columnwidth, keepaspectratio]{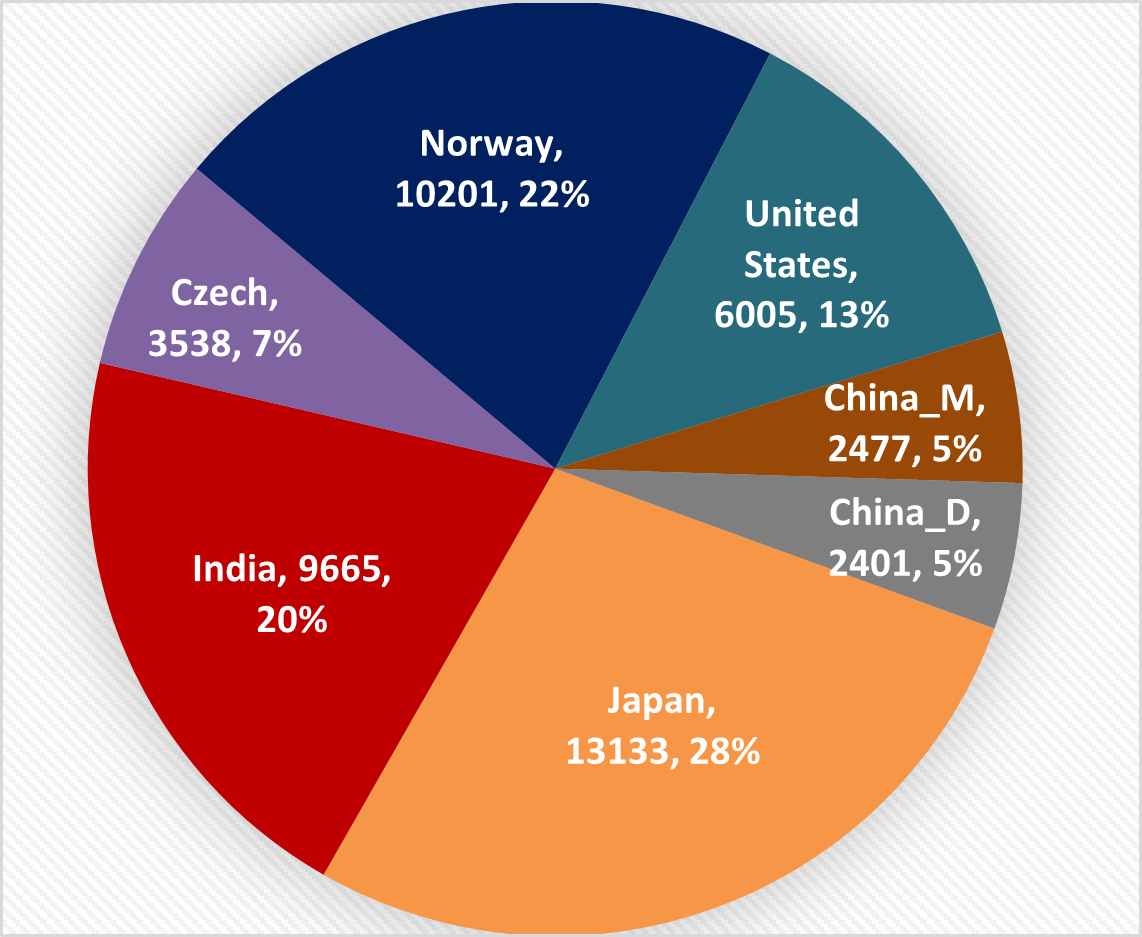}
\caption{Distribution of images across different countries in RDD2022}
\label{fig:Figure1_Statistics}
\end{figure}

\begin{figure}[!ht]
\centering
\includegraphics[width=1.0\columnwidth, keepaspectratio]{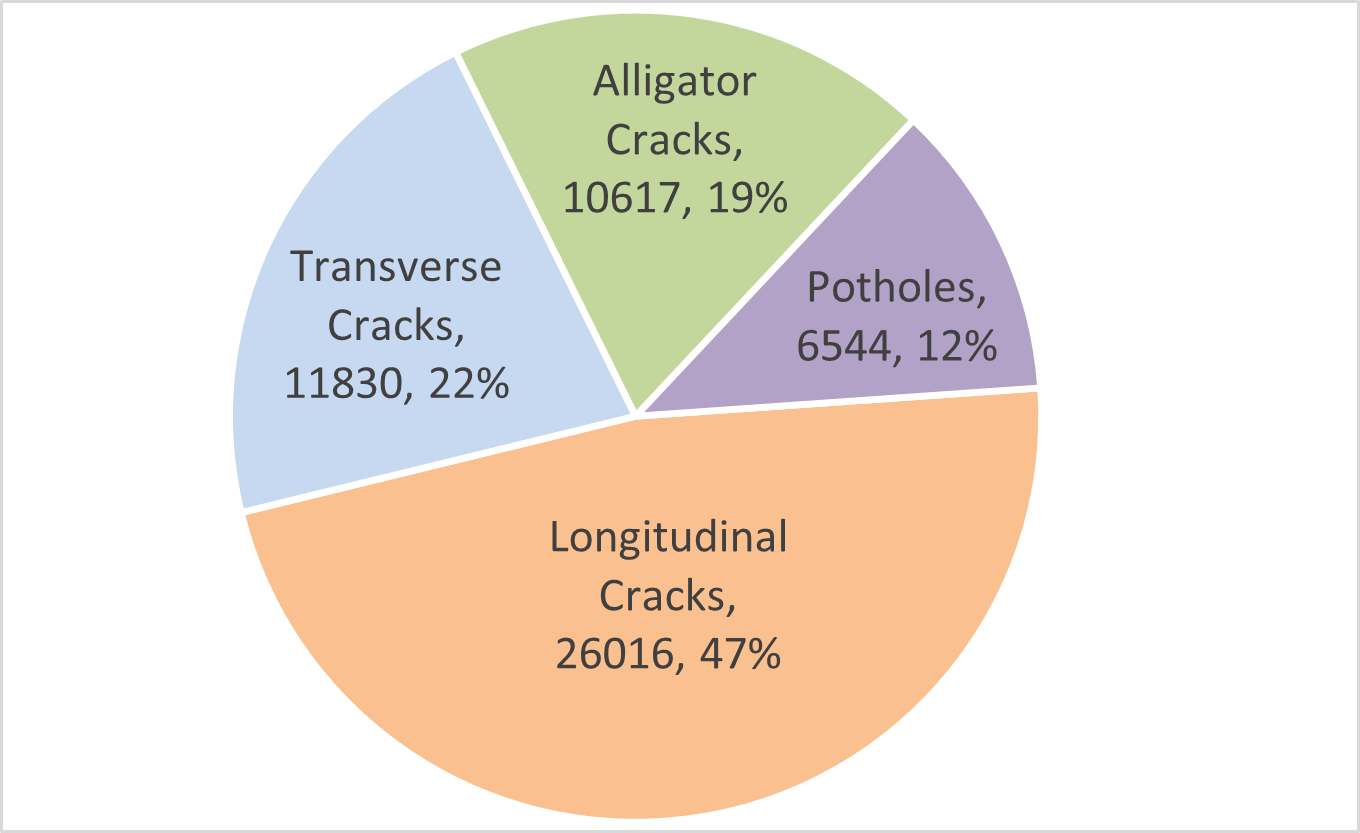}
\caption{Distribution of different damage type instances in training data released for CRDDC'2022}
\label{fig:Figure2_DamageTypes}
\end{figure}

\subsection{Tasks}
The challenge was organized in four phases with different tasks for the participants. The details are as follows.

\begin{enumerate}

\item \textbf{Phase 1: Data Solicitation}: The participants were asked to submit a report of their proposed road damage dataset (cracks, potholes, or any other damage categories) by providing the details through the corresponding questionnaire. The following categories of participants were considered:

\begin{enumerate}

\item Data Contributors (Data owners interested in contributing their data to CRDDC)
    
\item Information Providers (Data owners interested to participate by providing only the information related to their dataset, keeping the data private)
    
\item Data Recommenders (Researchers, Academicians, Freelancers, interested in recommending the inclusion of others' datasets to CRDDC)

\end{enumerate}

\item \textbf{Phase 2: Data Shortlisting}: Based on the reports submitted in phase 1, the shortlisted teams were asked to submit a link to their proposed dataset in Phase 2. After verification, the winners (data contributors) of phase 1 were announced. The solicited datasets were added to the Challenge datasets for the next phase of the challenge.
    
\item \textbf{Phase 3: Main Task}: The participants were asked to propose road damage detection models targeting the following 5 cases using any combination of the CRDDC dataset:

\begin{enumerate}
    \item Models applicable to all six countries (Overall category)
    \item Models applicable to India
    \item Models applicable to Japan
    \item Models applicable to the United States
    \item Models applicable to Norway
    
\end{enumerate}

The participants were allowed to propose models targeting one or multiple countries. For instance - Model A for Country 1, Model B for Country 2, Model C for Countries 3 and 4, or a single model for all countries could be proposed.

The LeaderBoard-1 (Overall category for all 6 countries) required the participants to propose an algorithm that automatically recognizes the road damages in an image captured from any of the underlying six countries. The recognition implies detecting the damage location in the image and identifying the damage type. Using this algorithm, the participants need to predict labels and the corresponding boundary boxes for damage instances present in each test data image. The predictions need to be submitted as a CSV file through the submission link provided on the website. The general rules to be followed while accomplishing the challenge tasks are stated below:	

\begin{enumerate}
    
\item Algorithm: Recently, Deep Learning models have become the first choice for applications involving object detection. However, the challenge allows all types of algorithms as long as the proposed solution can outperform other solutions.  
    
\item	Pre-trained Models: The challenge permits the use of transfer learning utilizing pre-trained models.
    
\item	Data: Participants are restricted to train their algorithms on the provided Road Data train sets. Collecting additional data for the target attribute labels is not allowed. However, it is permitted to increase training images using data augmentation, GANs, etc., artificially.
    
\item	License: The dataset released for the challenge is publicly available under the Creative Commons Attribution-ShareAlike 4.0 International License (CC BY-SA 4.0).
\end{enumerate}

\item \textbf{Phase 4: Report and Source Code}: Phase 4 required the participants to submit a report of their proposed solution along with the source code. 

\end{enumerate}

The following section provides details about the submission and evaluation of the results.

%% file: Submission_and_Evaluation.tex
\section{Submission and Evaluation}

\subsection{Submission of Results }

The running of the challenge consisted of four phases with submission at multiple stages, as described below.

\begin{enumerate}

	\item At the start of the challenge, the task for Phase 1 and Phase 2 were announced, and participants were asked to submit the information of the dataset they wanted to include for the $CRDDC$-Phase 3. The information was collected through a set of questionnaires released as Google Forms and the submission of a report for the proposed dataset. 
	
   \item In phase 2, the shortlisted participants of Phase 1 were asked to provide a link to their proposed dataset. Both images and annotations were required at this stage. 
	
   \item In phase 3, the $CRDDC$ dataset, RDD2022, comprising RDD2020 extended with the contribution of Phase 1 winners, was released. The dataset details are provided in our companion paper \cite{RDD2022_data}. Unlike $GRDDC$’2020, both Train and Test sets were released at once in $CRDDC$. The Train set included images and annotations, whereas the Test set included only images with no annotation files. 
   Participants were required to train their models using the Train set, infer the results for Test data and submit results to an evaluation server. The server was designed with an evaluation script, which calculates a score corresponding to every submission. The calculated score is added to the private leaderboard of the participants. Simultaneously, a public leaderboard is maintained on the challenge website ({\color{blue}\url{https://crddc2022.sekilab.global/}}) to report the highest score achieved by the teams on their private leaderboards. Five leaderboards for each category (private and public) were maintained to cover the five target cases mentioned in Phase 3 tasks.

   \item In the final phase, the participants were required to submit their source code to verify the replicability of the proposed algorithms.

\end{enumerate}	

The evaluation procedure for the submissions is described in the following sub-section.

\subsection{Evaluation}
The evaluation procedure for $CRDDC$’2022 is same as that of $GRDDC$’2020, except that there are five concurrent leaderboards for $CRDDC$, unlike $GRDDC$, which had two leaderboards for different test sets. The results submitted by the participants are evaluated using an evaluation script embedded in the server. The script utilizes two inputs: (i) the ground truth information of the test dataset, the annotations, and (ii) the file containing the participants' predictions. A prediction is considered correct if it satisfies the following two criteria:
	The area of overlap between the predicted bounding box and ground truth bounding box exceeds 50$\%$. The condition is described as $IoU > 0.5$. $IoU$ denotes Intersection Over Union. For more details regarding this, please refer to ~\cite{arya2020transfer}.
	The predicted label matches the actual label, as specified in the annotation file (ground truth) of the image. 
The script compares the two input files and calculates F1-Score for the submission. The F1-score is calculated as the Harmonic Mean of Precision and Recall values. Precision is the ratio of true positives to all predicted positive. The recall is the ratio of true positives to all actual positives. The details of the parameters are given below:

\begin{enumerate}
    \item  \textit{True Positive (TP)}: When a damage instance is present in the ground truth, and the label and the bounding box of the instance are correctly predicted with $IoU > 0.5$.
	 \item  \textit{False Positive (FP)}:  When the model predicts a damage instance at a particular location in the image, but the instance is not present in the ground truth for the image. This also covers the case when the predicted label doesn't match with the actual label.
	 \item \textit{False Negative (FN)}: When a damage instance is present in the ground truth, but the model fails to predict either the correct label or the bounding box of the instance. 

 \item \textit{Recall}:
\begin{equation}\label{eq1}
Recall = {\frac {(TP)}{(TP+FN)}}
\end{equation}

\item \textit{Precision}:
\begin{equation}\label{eq2}
Precision = {\frac {(TP)}{(TP+FP)}}
\end{equation}

\item \textit{F1-Score}:

\begin{equation}\label{eq3}
F1-Score = 2 \times {\frac {(Precision \times Recall)}{(Precision+Recall)}}
\end{equation}

\end{enumerate}

The F1 metric weights recall and precision equally. Thus, the competition favors moderately good performance on both over the outstanding performance on one and poor performance on the other. F1 is calculated separately for each of the five target cases defined in Phase 3 tasks. Finally, the average of all five scores is used to rank the teams. 

%% file: Results_and_Ranking.tex
\section{CRDDC'2022: Results and Rankings}

\begin{table*}[]
\caption{Details of the winners of Data Contribution Phase for CRDDC’2022 (Phase 1 and 2)}
\label{tab:Table_1_Data_Contributors}

\begin{tabular}{|c|m{2 cm}|m{6 cm}|m{4cm}|c|}

\hline
\centering \textbf{Rank} & \centering \textbf{Team Name}   & \centering \textbf{Contributors}                                                                                             & \centering \textbf{Affiliation} &  \textbf{Contribution}   \\\hline
1             & \centering NTNU                 & Madhavendra Sharma, Mamoona Birkhez   Shami, Muneer Al-Hammadi, Alex Klein-Paste, Helge Mork, and Frank Lindseth, & Norwegian University   of Science and Technology, Norway & Data from Norway                     \\\hline
2             & \centering AGRD                 & Van Vung Pham, Du   Nguyen                                                                                        & Sam Houston State University, United States              & Data from the United States          \\ \hline
3             & \centering Futuristic Detection & Jingtao Zhong,   Hanglin Cheng, Jing Zhang                                                                        & Southeast   University, China     & Data from China \\ \hline
\end{tabular}
\end{table*}


\begin{table*}[]
\caption{CRDDC Ranks and Scores of top 11 teams (Phase 3 and 4)}
\label{tab:Table_2_Ranks_and_Scores}

\begin{tabular}{|c|c|c|c|c|c|c|c|}
\hline
\multicolumn{1}{|c|}{\textbf{Rank}} & \multicolumn{1}{c|}{\textbf{Team Name}} & \multicolumn{1}{c|}{\begin{tabular}[c]{@{}c@{}}\textbf{LeaderBoard - 1} \\ (\textbf{overall 6 countries})\end{tabular}} & \multicolumn{1}{c|}{\begin{tabular}[c]{@{}c@{}}\textbf{LeaderBoard - 2}\\ (\textbf{India})\end{tabular}} & \multicolumn{1}{c|}{\begin{tabular}[c]{@{}c@{}}\textbf{LeaderBoard - 3}\\ (\textbf{Japan})\end{tabular}} & \multicolumn{1}{c|}{\begin{tabular}[c]{@{}c@{}}\textbf{LeaderBoard - 4}\\ (\textbf{Norway})\end{tabular}} & \multicolumn{1}{c|}{\begin{tabular}[c]{@{}c@{}}\textbf{LeaderBoard - 5}\\ (\textbf{United States})\end{tabular}} & \multicolumn{1}{c|}{\textbf{Average Score}} \\ \hline
1                          & ShiYu\_SeaView                 & 0.770                                                                                            & 0.583                                                                                     & 0.789                                                                                     & 0.595                                                                                      & 0.844                                                                                             & 0.716                              \\ \hline
2                          & DongjunJeong                   & 0.743                                                                                            & 0.540                                                                                     & 0.750                                                                                     & 0.538                                                                                      & 0.801                                                                                             & 0.674                              \\ \hline
3                          & MDPT                           & 0.741                                                                                            & 0.516                                                                                     & 0.735                                                                                     & 0.504                                                                                      & 0.817                                                                                             & 0.663                              \\ \hline
4                          & SGG-RS-Group                   & 0.727                                                                                            & 0.545                                                                                     & 0.727                                                                                     & 0.481                                                                                      & 0.779                                                                                             & 0.652                              \\ \hline
5                          & IRCV-URV                       & 0.726                                                                                            & 0.518                                                                                     & 0.735                                                                                     & 0.498                                                                                      & 0.779                                                                                             & 0.651                              \\ \hline
6                          & IMSC                           & 0.728                                                                                            & 0.542                                                                                     & 0.725                                                                                     & 0.478                                                                                      & 0.774                                                                                             & 0.649                              \\ \hline
7                          & NJUPT                          & 0.718                                                                                            & 0.559                                                                                     & 0.720                                                                                     & 0.458                                                                                      & 0.743                                                                                             & 0.640                              \\ \hline
8                          & TUT                            & 0.694                                                                                            & 0.519                                                                                     & 0.773                                                                                     & 0.464                                                                                      & 0.727                                                                                             & 0.636                              \\ \hline
9                          & MILA                           & 0.697                                                                                            & 0.494                                                                                     & 0.716                                                                                     & 0.462                                                                                      & 0.775                                                                                             & 0.629                              \\ \hline
10                         & SIAI                           & 0.612                                                                                            & 0.417                                                                                     & 0.608                                                                                     & 0.424                                                                                      & 0.663                                                                                             & 0.545                              \\ \hline
11                         & kubapok                        & 0.603                                                                                            & 0.421                                                                                     & 0.594                                                                                     & 0.383                                                                                      & 0.649                                                                                             & 0.530                              \\ \hline
\end{tabular}
\end{table*}



\begin{table*}[]
\centering
\caption{Details of the Contributors (Top 11 teams for CRDDC’2022 - Phase 3 and 4)}
\label{tab:Table_3_Contributors}
\begin{tabular}{|c|m{5.5cm}|m{5cm}|c|}

\hline
\textbf{Team Name} & \centering \textbf{Contributors}                                                                         & \centering \textbf{Affiliation}                                                                                                          & \textbf{Country}   \\ \hline
ShiYu\_SeaView     & Wenchao Ding, Xu Zhao,   Bingke Zhu, Yinglong Du, Guibo Zhu, Tao Yu, Lei Li, and Jinqiao Wang & Objecteye Inc.                                                                                                                & China              \\ \hline
DongjunJeong       & (1) Dongjun Jeong, ((2) Jua   Kim                                                             & (1) OPGG, (2) X (freelancer)                                                                                                  & Republic of Korea  \\ \hline
MDPT               & Vung Pham, Du Nguyen,   Christopher Donan                                                     & Sam Houston State University                                                                                                  & United States      \\ \hline
SGG-RS-Group       & Yao Tang, Xusi Liao, Jiang   He, Haoliang Feng, Hongzan Jiao, Xin Su, Qiangqiang Yuan         & Wuhan University                                                                                                              & China              \\ \hline
IRCV-URV           & Ammar M. Okran, Mohamed   Abdel-Nasser, Hatem A. Rashwan, Domenec Puig                        & Universitat Rovira i Virgili                                                                                                  & Spain              \\ \hline
IMSC               & Seon Ho Kim, Maitry Bhavsar,   Utkarsh Baranwal, Abdullah Alfarrarjeh                         & University of Southern California (USA), Indian   Institute of Technology Patna (India), German Jordanian University (Jordan) & USA, India, Jordan \\ \hline
NJUPT              & Shihao Han, Guoping Jiang, Yingjiang   Zhou, Haowen Xu, Jiajing Ying, Huan Yang                    & Nanjing University of Posts   and Telecommunications, College of Automation and Artificial Intelligence                       & China              \\ \hline
TUT                & Lu Yang, Hualin He, Tao Liu                                                                   & Tianjin University of   Technology                                                                                            & China              \\ \hline
MILA               & Poonam Kumari Saha,   Yoshihide Sekimoto                                                      & Sekimoto Lab, The University   of Tokyo                                                                                       & Japan              \\ \hline
SIAI               & Jeon Yun Tae, Dai Quoc Tran                                                                   & Department of Global Smart   City, Sungkyunkwan University                                                                    & Korea              \\ \hline
kubapok            & Jakub Pokrywka                                                                                & Adam Mickiewicz University                                                                                                    & Poland             \\ \hline
\end{tabular}

\end{table*}


\begin{table*}
\centering
\caption{Details of the solutions proposed by top 11 teams for CRDDC’2022 - Phase 3 and 4}
\label{tab:Table_4_Proposed_Solutions_CRDDC}
\begin{tabular}{|c|m{1.6 cm}|m{3.3 cm}|c|m{6.8 cm}|}

\hline
\textbf{Team Name}    & \centering \textbf{Data utilized}                                            & \centering \textbf{Proposed Solution}                                                               & \textbf{Ensemble learning} & \textbf{Main Feature}                                                                                                                     \\ \hline
ShiYu\_SeaView    & RDD2022  & Ensemble model based on YOLO-series and Faster RCNN-series models                     &  Yes         & Multi-model ensemble: 1.YOLO-series models to achieve Dense predictions, and 2. the   Faster RCNN-series model with Swin Transformer as backbone to achieve more accurate predictions by their large receptive fields and two-stage methods.         \\ \hline
DongjunJeong             & RDD2022                                                                 & YOLOv5x P5 and P6 Ensemble with Image patch                                                 &  Yes         & 1. Use of Image patch strategy to deal with high-resolution images. and 2. Use of pre-trained weights based on GRDDC'2020 winning solution. \\ \hline
MDPT                    & RDD2022                                                                 & YOLOv7 with train and test image augmentations,label smoothing, and coordinate attentions & No      & 1. Cropping the Norway images to focus only on road areas and support training with limited hardware resources. 2. Training country-specific models using data from all countries by varying the   images used for validation.   \\ \hline
IMSC                                        & RDD2022                                                                                  & YOLOv5 based ensemble                                                                                         &  Yes                          & Use of customized Anchor Boxes tuned according to the used training dataset to fit the sizes of objects of interest.                                                  \\ \hline
SGG-RS-Group                                & RDD2022 except China\_Drone                                                              & The ensemble model based on   YOLOv5 and attention modules                                                    &  Yes                          & Use of Attention modules to enhance important feature maps, such as highlighting road regions in the image.                                                           \\ \hline
IRCV-URV                                    & RDD2022                                                                                  & Yolov7-based ensemble                                                                                         &  Yes                          & Experiments to ensemble models trained with different image-size and architectures, including the use of Atrous Spatial Pyramid Pooling.                              \\ \hline
NJUPT               & RDD2022 except China\_Drone                                         & yolov7x, yolov5x, yolov5x-transformer                                                                      &  Yes                          & Experiments to ensemble models based on the input image size, attention mechanisms, and data augmentations. Shows that more models don’t always mean higher accuracy. \\ \hline

MILA & RDD2022 except China\_Drone (Only positive images) & YOLOv7-based ensemble  &  Yes        & 1. Innovative Data Augmentation based on 17 techniques. 2. Consideration of only Positive samples to support faster training through image caching.                                                                           \\ \hline
TUT                  & RDD2022                                                             & YPLNet (Yolov5s + Pyramid   Squeeze Attention + Large-field Contextual Feature Integration)                   & No                          & Extraction and fusion of multi-scale contextual features using PSA and LCFI with limited computational cost.                                                          \\ \hline
SIAI & RDD2022 except China\_Drone                      & YOLOv5 + VFNet in MMDetection                                                               &  Yes        & 1. Image scaling applied to   Norway images. 2.   VFNet worked well with rectangular images and Yolov5 for square.                                                                                                     \\ \hline
kubapok             & RDD2022                                                           & YOLOv5 based ensemble                                                                                         &  Yes                          & Optimal GPU time usage by an ensemble of models trained with multiple levels of augmentations.                                                                          \\ \hline
\end{tabular}

\end{table*}

More than 50 teams from 19 different nationalities registered for the challenge. The related data is available on the challenge website ({\color{blue}\url{https://crddc2022.sekilab.global/participants/}}). The details of shortlisted solutions and winners of different phases are provided as follows.

\subsection{Phase 1 and 2: Data Solicitation}
For the data contribution phase, the datasets submitted by the three teams listed in Table ~\ref{tab:Table_1_Data_Contributors}, were found the most suitable for merging with the existing RDD2020 dataset. The combined dataset was released as RDD2022 and provided to Phase 3 participants. The RDD2022 data has been made freely available at the FigShare repository \cite{RDD2022_data}. The details are provided in \cite{RDD2022_paper}.

\subsection{Phase 3 and 4: Model Contribution}

The top 11 teams were shortlisted based on the average F1 score achieved by the proposed solutions, the submissions of source code, and a detailed report. Table  \ref{tab:Table_2_Ranks_and_Scores} lists the ranking achieved by these teams and their scores corresponding to the five leaderboards. The details of the contributors are presented in tables \ref{tab:Table_3_Contributors}. The top 5 teams achieved an average F1 score of more than 65\% and proposed efficient methods using ensemble learning and several data augmentation techniques. Table \ref{tab:Table_4_Proposed_Solutions_CRDDC} summarizes the details of the top 11 solutions proposed targeting the six countries together (Overall\_6\_countries leaderboard).

It may be noted that, the state-of-the-art object detectors are grouped into two categories: two-stage detector\cite{fast, faster, mask, cascade} and one-stage detector \cite{ yolov1, yolov2, yolov3, bochkovskiy2020yolov4, lin2017focal, atss, yolov5_software_link, yolov7}. As the name suggests, the two-stage detector works in two steps: it first extract proposals and then send them into the later network for fine-grained classification and location. Whereas, the one-stage detector performs these two tasks in one step - the images are fed into the network to get the result directly. As a result, one-stage detectors are usually faster. 

The winning method for CRDDC'2022, proposed by team ShiYu\_SeaView, is based on an ensemble of one-stage and two-stage detectors. For one-stage, the team adopted YOLOv5\cite{yolov5_software_link} and YOLOv7\cite{yolov7}. YOLOv5\cite{YOLOv5} is choosen because it showed high performance in solutions proposed through GRDDC'2020\cite{GRDDC2020} targeting the same domain of road damage detection. The YOLOv7\cite{yolov7} is selected because it has several trainable bag-of-freebies, e.g., model re-parametrization, and auxiliary head, and is a new state-of-the-art real-time object detector.

Likewise, the team adopted Cascade RCNN\cite{cascade} models for two-stage detectors after analyzing the top-ranking solutions\cite{2020rank1,rdd2020rank2, 2020rank3} for GRDDC'2020\cite{GRDDC2020}. From the anlaysis, the team inferred that YOLO series models perfer to have a higher recall rate, while Cascade RCNN series models prefer to have a higher precision rate and works better for small damage. Based on the inference, ensemble learning of these two types of models was proposed to have complementary positive effects on the performance. 

Further, the team also utilized Swin Transformer\cite{swin, swinv2} as the Backbone and Deformable ROI Pooling as the ROI head of the two-stage detector. This is because the self-attention module (transformer\cite{attention}) enables better extraction of semantic information and has a larger receptive field. That is considered useful for the current problem domain because cracks often have very long and narrow morphologies and traditional $3\times3$ convolution kernels may not extract these features properly. Deformable convolution\cite{dai2017deformable, zhu2019deformable} sets the learnable offsets to solve geometric transformation problems. Besides, the ROI Pooling\cite{dai2017deformable} can better adapt to the special shape of road damage.

Similar to the Rank 1 team, ShiYu\_SeaView, the team ranked second in CRDDC, Dongjun Jeong, utilized the GRDDC'2020 winning solutions, to decide their approach for CRDDC'2022. The proposed approach include the use of transfer learning using the COCO dataset pre-trained weight and the pre-trained weight of GRDDC’2020 winner USC-InfoLab (Team IMSC \cite{2020rank1}).

This shows the contribution made by the road damage detection challenges and participants to state-of-the-art for the domain over the years.

%% file: Discussion_and_FutureScope.tex
\section{Discussion and Future Scope}

\begin{table*}[]
\caption{Comparison of Road Damage Detection Challenges from 2018 to 2022}
\label{tab:Table_5_RDD_Comparison}

\begin{tabular}{|m{2.4 cm}|m{3.5 cm}|m{3.5 cm}|m{7 cm}|c|}
\hline
 &
  \textbf{RDDC'2018} &
  \textbf{GRDDC'2020} &
  \textbf{CRDDC'2022} \\ \hline
  
\textbf{\#Countries targeted} &
  1 &
  3 &
  6 \\ \hline
  
\textbf{\#Teams} &
  59 &
  121 &
  64 (at the time of writing) \\ \hline
  
\textbf{Difficulty level} &
  High (First challenge of the series) &
  Higher (mainly due to the geographical diversity of target countries) &
  Highest (Geographical diversity + multi-resolution images), multiple sub-tasks \\ \hline
  
\textbf{Main task} &
  Model for only Japan &
  Combined model for 3 countries &
  More challenging, combined model for 6 countries, an individual model for 4 countries \\ \hline
  
\textbf{Damage categories} &
  8 &
  4 &
  4 \\ \hline
  
\textbf{Positive points} &
  (i) Standard Method of Assessment, (ii) Fixing the test data for evaluating the performance &
  (i) Foundation for proposing multi-country solutions, (ii) Availability of heterogeneous data to train more robust models. &
  (i) Heterogeneity in the data improved - multi-sensor, multi-country, multi-resolution images, (ii) Data contribution by participants allowed, (iii) Standard Method of Assessment for multiple target use cases, (iv) More than 5500 labels provided for each class. \\ \hline
  
\textbf{Type of challenge} &
  Model-centric, but participants allowed to augment the data artificially. &
  Model-centric, but participants allowed to augment the data artificially. &
  Model-centric, with more freedom for the participants to experiment with the data.  Artificial augmentation - allowed. Additionally, Phase 1 invited participants to contribute their data. \\ \hline
  
\textbf{Test data} &
  Two sets - Test 1 and Test2, released at different times, comprising images from Japan. &
  Two sets - Test 1 and Test2, released at different times, comprising images from Japan, India, and Czech Republic. &
  One set - Test, comprising images from underlying 6 countries, released at once together with training data. \\ \hline
  
\textbf{Time for training the models} &
  Around 4 months before the test1 deadline &
  Around 4 months before the test1 deadline &
  Around 2 months \\ \hline
  
\textbf{Number of leaderboards} &
  Two with different deadlines (Test 1 and Test2) &
  Two with different deadlines (Test 1 and Test2) &
  Five with same deadline (Overall\_6\_countries, India, Japan, Norway, United States) \\ \hline
  
\textbf{Ranking Criteria} &
  Average score for test1 and test2 &
  Average score for test1 and test2 &
  Average score for all 5 leaderboards \\ \hline
  
\textbf{Best score achieved} &
  0.665 (for  Japan) &
  0.68 (for India, Japan, and Czech Republic) &
  0.716 (average for 5 leaderboards), 0.769 (for all 6 countries combined), 0.780 (for Japan) \\ \hline
  
\textbf{Best performing model} &
  Ensemble of Faster-RCNN and SSD, with ImageNet pre-trained ResNet-101 and VGG-16 for the backbone models, respectively &
  YOLOv5-based ensemble model &
  Ensemble model based on YOLO-series and Faster RCNN-series models \\ \hline
\end{tabular}

\end{table*}

The significance of Road Damage Detection challenges lies in providing a rich dataset and a standardized evaluation framework for comparing different methods. The CRDD challenge has lived up to the mark of these expectations. Table 5 presents a comparative summary of the three road damage detection challenges organized till now concerning multiple factors. Clearly, CRDDC’2022 made significant contributions to the existing literature by consideration of multiple use cases. Following are some directions that may be considered for future road damage detection challenges.

\begin{enumerate}[]
\item Due to the consideration of multiple phases in CRDDC’2022, the participants got lesser time to perform experiments and train the models (Phase 3). Some participants requested to reopen the leaderboards. Even though new leaderboards on the challenge website were created to support more experiments by CRDDC participants and users who couldn’t join the competition, future challenges may plan and allocate more time for the competition tasks.

\item	Inclusion of high-resolution images in the dataset RDD2022, released through CRDDC'2022, led to the requirement of heavy-computation resources to process the data. This restricts the participation of users with limited resources. Future challenges may address this issue by providing support for computational resources, such as Graphics Processing Units (GPU), to participants in need.

\item	Though CRDDC allowed participants to propose and contribute the data of their choice in Phase 1, the participants were restricted to utilizing only the provided training data in Phase III. In the future, the utilization of other data sources may also be allowed.

\item Until now, $RDD$ challenges focussed only on the detection and classification of road damages. Future challenges may introduce new tasks such as Severity analysis, Pixel-level damage analysis for covering a wider spectrum for road condition monitoring.

\item Further, multiple evaluation metrics can be introduced for assessing the performance better. For instance, considering the inference speed and size of the proposed model in addition to F1-Score would be useful for developing real-time smartphone-based object detection methods.

\item Furthermore, data balancing at multiple levels may be considered for the upcoming years concerning countries, damage categories, etc.

\item	Additionally, the complexity level for CRDDC was higher than that of previous challenges on various factors listed in Table \ref{tab:Table_5_RDD_Comparison}. Simpler challenges may be considered in the future to analyze the performance of the same model for different countries, keeping the data collection process (mechanism, image capturing device, image resolution, etc.) uniform.

\end{enumerate}

%% file: conclusion.tex
\section{Conclusion}
\label{sec:con}

The IEEE BigData cup, Crowdsensing-based Road Damage Detection challenge, focuses on the detection and classification of road damage in different parts of the world, particularly India, Japan, the Czech Republic, Norway, the United States, and China. It includes four common damage categories: Longitudinal Cracks, Transverse Cracks, Alligator Cracks, and Potholes. Since the data comes from different regions of the world, including developed and developing countries, the quality of road hardening varies. Further, the degree of damage, area, aspect ratio, etc., of each type of damage differs significantly. Furthermore, the data has been captured using multiple mechanisms – including Smartphone-mounted vehicles, wide-angle cameras, drones, motorbikes etc., with different image resolutions. Therefore, the images have high levels of variation in intensity and texture content. Thus, the BigData cup offers a major challenge to the participants to explore the most relevant techniques.

This paper summarizes the challenge, elaborating on the main characteristics, the complexities of the tasks to solve, and the overall evaluation methodology. It also describes the main features of the approaches proposed by the highest-ranked contestants and proposes recommendations for future challenges. In a nutshell, the contest and the summary paper may be used as a benchmark for the state of the art since multiple solutions are presented with valuable insights and contribution from the research community, and the dataset is open for future testing.

%% file: acknowledgment.tex
\section*{Acknowledgment}

We thank all the participants of the challenge for their contributions. We also thank the sponsor UrbanX Technologies for providing the requisite funds.